\documentclass{article}

\usepackage{booktabs}
\usepackage{adjustbox}
\usepackage{wrapfig}

\usepackage[preprint]{neurips_2024_style}
\setcitestyle{numbers}

\usepackage{microtype}
\usepackage{graphicx}
\usepackage{subfigure}
\usepackage{caption}
\usepackage{subcaption}
\usepackage{float}
\usepackage{booktabs} 

\usepackage[utf8]{inputenc} 
\usepackage[T1]{fontenc}    
\usepackage{hyperref}       
\usepackage{url}            
\usepackage{booktabs}       
\usepackage{amsfonts}       
\usepackage{nicefrac}       
\usepackage{microtype}      
\usepackage{xcolor}         
\usepackage{amsmath}
\usepackage{amsthm}
\usepackage{amssymb}
\usepackage{mathrsfs}
\usepackage{MnSymbol}
\usepackage{wrapfig}
\newtheorem{definition}{Definition}

\newtheorem{proposition}{Proposition}
\usepackage{multirow}
\newtheorem{theorem}{Theorem}

\newtheorem{corollary}{Corollary}

\title{Revisiting Spurious Correlation in Domain Generalization}

\author{%
  Bin Qin\thanks{Contributed equally to this work.}, Jiangmeng Li$^*$\thanks{Corresponding author.}, Yi Li, Xuesong Wu, Yupeng Wang, Wenwen Qiang, Jianwen Cao$^\dagger$\\
  Institute of Software\\
  Chinese Academy of Sciences\\
  Beijing, China \\
  \texttt{\{jiangmeng2019,jianwen\}@iscas.ac.cn} \\
}

\begin{document}

\maketitle

\begin{abstract}
Without loss of generality, existing machine learning techniques may learn \textit{spurious correlation} dependent on the domain, which exacerbates the generalization of models in out-of-distribution (OOD) scenarios. To address this issue, recent works build a structural causal model (SCM) to describe the causality within \textit{data generation process}, thereby motivating methods to avoid the learning of spurious correlation by models. However, from the machine learning viewpoint, such a theoretical analysis omits the nuanced difference between the data generation process and representation learning process, resulting in that the causal analysis based on the former cannot well adapt to the latter. To this end, we explore to build a SCM for \textit{representation learning process} and further conduct a thorough analysis of the mechanisms underlying spurious correlation. We underscore that adjusting erroneous covariates introduces bias, thus necessitating the correct selection of spurious correlation mechanisms based on practical application scenarios. In this regard, we substantiate the correctness of the proposed SCM and further propose to control confounding bias in OOD generalization by introducing a propensity score weighted estimator, which can be integrated into any existing OOD method as a plug-and-play module. The empirical results comprehensively demonstrate the effectiveness of our  method on synthetic and large-scale real OOD datasets.
\end{abstract}

\section{Introduction}
Machine learning techniques excel at finding correlations between the input data and the task-related labels in various real-world applications \cite{liu2021towards,vapnik1998statistical}. However, conventional approaches perform well towards in-distribution (ID) scenarios, i.e., the target data are i.i.d. with the \textit{seen} training set data, while falls short of generalizing to out-of-distribution (OOD) scenarios, encompassing the \textit{unseen} data obeying a distribution that is heterogeneous with the seen training set distribution. To understand the intrinsic mechanism behind such an issue, recent works~\cite{deng2024robust,chen2024understanding} explore that canonical learning paradigm unavoidably leads the model to learn the \textit{spurious correlation} between the \textit{domain-dependent} feature of input data and the task-related label, resulting in that the inconsistency of OOD domains widely degenerates the generalization performance of the model. For instance, typical machine learning models overly rely on spurious correlations~\cite{geirhos2020shortcut,he2021towards} between background or minor objects and labels to exploit shortcuts for prediction in ID scenarios, e.g., camels stand on deserts, metal markers appear on specific positions of chest X-ray scans~\cite{beery2018recognition,kirichenko2023last}.

State-of-the-art methods dedicate to mitigate or eliminate the undesired spurious correlation by exploring the latent causal mechanism~\cite{zhang2021causaladv,lu2021invariant,liu2021learning,peters2016causal} of machine learning paradigms and further introducing practical approaches to learn the invariance of causal features in OOD scenarios~\cite{ahuja2021invariance,lin2022zin,yang2024invariant}. Concretely, such methods primarily build a structural causal model (SCM) \cite{pearl1998graphs,pearl2012causal} behind the \textit{data generation process}, which divides the feature contained by the input data into two separated parts: the \textit{invariant} feature involving the domain-agnostic information and the \textit{spurious} feature involving domain-dependent information, and the latter incurs in a certain spurious correlation. Inspired by the causal analysis, benchmark methods aim to find invariant features by introducing new loss functions or regularization designs that incorporate specific invariance restrictions across variant domains into the representation learning process~\cite{krueger2021out,koyama2020invariance,mroueh2021fair}. However, there exists a focal issue challenging the improvement of such methods stemming from the deficiency of solid and thorough causal analysis. In this regard, for the construction of SCM \cite{lu2021invariant,fan2022debiasing}, existing methods unexpectedly omit the nuanced difference between the data generation process and representation learning process, leading to that the causal analysis based on the former cannot well adapt to the latter. For the implementation of SCM-based adjustment \cite{mao2021generative}, the mechanisms underlying spurious correlation require scrutinizing, since adjusting erroneous covariates introduces bias, thus necessitating the correct selection of spurious correlation mechanisms based on practical application scenarios.

To this end, we propose to develop a novel SCM for representation learning process, and further demonstrate that the candidate spurious correlation mechanisms include: 1) \textit{fork-specific} spurious correlation caused by latent common cause; 2) \textit{collider-specific} spurious correlation caused by conditional common effect. For the proposed SCM, we substantiate the correctness of collider-specific spurious correlation based on OOD applications. Hence, motivated by the causal analysis, we propose a practical approach to control confounding bias in OOD generalization problems. Concretely, we introduce a propensity score weighted (PSW) estimator to rescale the observed distribution, simulating sampling from the post-intervention distribution without extra experiments. The PSW regularization term can be integrated into any existing OOD method as a plug-and-play module, and our experimental results on synthetic and large-scale real datasets empirically demonstrate that the PSW regularizer can enhance the performance of various state-of-the-art OOD methods.

Our contributions are as follows: 1) We propose a general SCM demonstrating the representation learning process for the domain generalization problem; 2) We provide a thorough analysis upon the consistency and inconsistency between spurious correlation mechanisms, and substantiate the correctness of collider-specific spurious correlation for OOD-oriented SCM; 3) We rigorously demonstrate that by expanding the set of available values for adjusted variables, it is guaranteed to identify the strength of the selection propensity of latent spurious features towards invariant features. This allows us to assign a propensity score weight to each observed sample, thereby reweighting the observational distribution; 4) The well-conducted empirical validation proves that our method can widely obtain consistent performance gains compared to baselines on sufficient benchmarks, encompassing PACS \cite{pacs}, OfficeHome \cite{officehome}, and etc.

\section{Spurious Correlation on OOD Generalization}
\label{headings}

\subsection{Problem Setting}
Assuming that a domain set, including different domains, is denoted as $\mathcal{D}=\left \{ D^{v} \right \}_{v=1}^{m}$, where $D^{v}=\left \{ \left ( \mathbf{x_{i}^{v}},y_{i}^{v} \right ) \right \}_{i=1}^{N_{v}}$ represents a domain indexed by $v\in V$, containing the collected data and corresponding labels. $N_{v}$ is the number of samples in domain $D_{v}$. Let $X^{v}=\left \{ \mathbf{x_{i}^{v}} \right \}$ be the non-empty input space, and $Y^{v}=\left \{ y_{i}^{v} \right \}$ be the true labels, where $\mathbf{x_{i}^{v}}\in X^{v}\subset \mathbb{R}^{^{d}}$, and $y_{i}^{v}\in Y^{v}\subset \mathbb{N}$. We assume that the samples in $D^{v}$ follow the distribution $\mathbb{P}^{v}\left ( X^v,Y^v \right )$, where generally $\mathbb{P}^{v}\left ( X^v,Y^v \right )\neq \mathbb{P}^{{v}'}\left ( X^{v'},Y^{v'} \right )$, indicating that the elements in $\mathcal{D}$ are out-of-distribution (OOD). Given $s$ training domains $\mathcal{D}_{train}=\left \{ D^{v}\in \mathcal{D}\mid v=1,2,...,s \right \}$, we aim to minimize prediction loss on unseen test domains $\mathcal{D}_{test}=\mathcal{D}\backslash \mathcal{D}_{train}$ by learning an \textbf{invariant function} $f:X\rightarrow Y$ on $\mathcal{D}_{train}$ as follows:
\begin{equation}
    \min_f\mathbb{E}_{(\mathbf{x},y)\in\mathcal{D}_{test}}[\ell(f(\mathbf{x}),y)].
\end{equation}

\subsection{Learning Representation Process}
\begin{figure}
    \centering
    \includegraphics[width=0.9\textwidth]{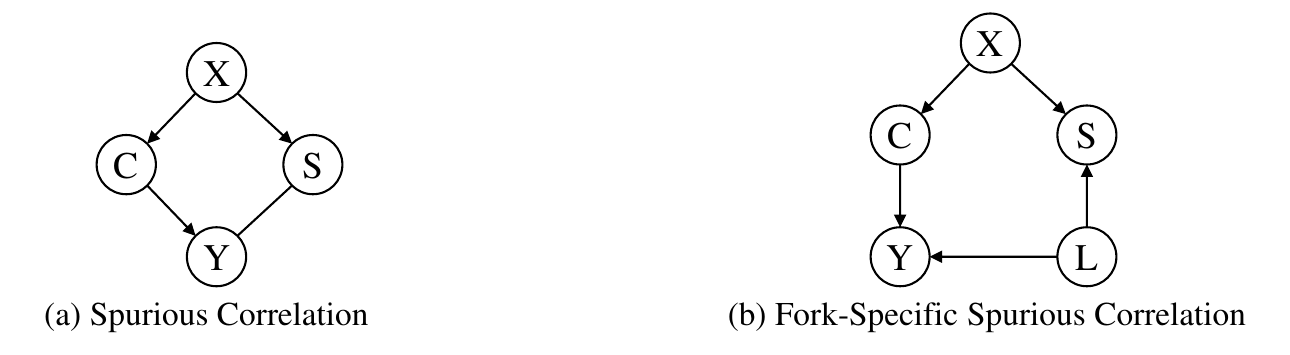}
    \caption{The SCM graph of representation learning for OOD. In Figure (a), the dash between $S$ and $Y$ represents the complex spurious correlation, where the direction of causal arrows cannot be determined. Figure (b) illustrates fork-specific spurious correlation, replacing $S - Y$ with $S\leftarrow L\rightarrow Y$, which is widely accepted as the latent common cause proposition in causal inference.}
    \label{fig:example1}
\end{figure}

We formally model the OOD generalization from the perspective of representation learning. Four variables, namely $X$, $C$, $S$, and $Y$, represent the dataset, invariant feature, spurious feature, and ground truth, respectively. We depict the SCM graph in Figure \ref{fig:example1}(a) based on the causal relationships between these variables. Each set of parent-child nodes in the directed acyclic graph (DAG) $G$ represents a deterministic function $x_{i}=f_{i}\left ( pa_{i},\epsilon _{i} \right )$, where $pa_{i}$ is the parent node of $x_i$ in $G$, and $\epsilon _{i}$ is a jointly independent, arbitrarily distributed random disturbance. In Appendix \ref{A}, we provide detailed causal background knowledge~\cite{pearl2009causality}.

We take the ColoredMNIST~\cite{arjovsky2019invariant} dataset as an example, and provide a detailed explanation:
\begin{enumerate}
    \item  $C\leftarrow X\rightarrow S$. This fork structure~\cite{pearl2016causal} indicates that invariant features $C$ (digit) and spurious features $S$ (color) are learned from the data $X$ (image). The variables $C$ and $S$ can be determined by functions $C=f_C(X,\epsilon _{C})$ and $S=f_S(X,\epsilon _{S})$, respectively.
    \item $C\rightarrow Y$. The invariant feature $C$ is the only explicit parent node of the ground truth label $Y$, implying that we can determine $Y=f_Y(C,\epsilon _{Y})$ through $C$.
    \item $S - Y$. The dash indicates a \textbf{correlation} between variables, without a clear \textbf{causation} determination, i.e., which variable is the parent is undetermined. This is because the causal relationship between spurious features and ground truth labels cannot be definitively established~\cite{liu2021learning}: $Y\rightarrow S$ possibly holds, for instance~\cite{ahuja2020empirical}, if we stipulate that small digits (less than or equal to 4) are associated with red color and large digits (greater than or equal to 5) with green color; $S\rightarrow Y$ possibly holds, as humans might use background color as an aid in determining labels when digits are too blurry to discern~\cite{ahuja2021invariance}, or random colors might be added to digit images. The mixing of data from different domains obscures the causal relationship between $S$ and $Y$, leading to a \textbf{spurious correlation} between $S$ and $Y$. Thus, we can derive the following theoretical definition:
\end{enumerate}

\begin{definition}
    \label{assumption 1} \textbf{(Spurious correlation~\cite{pearl2009causality})}
    Two variables, $X$ and $Y$, exhibit spurious correlation if they are dependent under certain conditions, and there exist two other variables ($Z_1$ and $Z_2$) and two scenarios ($S_1$ and $S_2$) such that:
    \begin{enumerate}
        \item Given $S_1$, $Z_1$ is dependent of $X$ ($Z_1 \nupmodels X \mid S_1$), and $Z_1$ is independent of $Y$ ($Z_1 \upmodels Y \mid S_1$).
        \item Given $S_2$, $Z_2$ is dependent of $Y$ ($Z_2 \nupmodels Y \mid S_2$), and $Z_2$ is independent of $X$ ($Z_2 \upmodels X \mid S_2$).
    \end{enumerate}
\end{definition}

For ease of understanding, we demonstrate the explanation of Definition \ref{assumption 1} by introducing Figure \ref{fig:example1}(a). Concretely, utilizing condition 1, given $S_1=C$ and $Z_1=X$, $X$ is dependent of $S$ but independent of $Y$, thus ruling out $S\rightarrow Y$. Using condition 2, given $S_2=X$ and $Z_2=C$, $C$ is dependent of $Y$ but independent of $S$, thus ruling out $Y\rightarrow S$, such that $S$ and $Y$ exhibits a spurious correlation.

\paragraph{Fork-Specific Spurious Correlation.}
We aim to learn domain-invariant representations, denoted as $C$, which are not confounded by spurious factors when identifying the causal effect $C\rightarrow Y$. Identification of the direct causal effect $C\rightarrow Y$ requires inference within a DAG, where the direction of arrows between variables must be explicitly known. However, the presence of spurious correlation $S - Y$ in Figure \ref{fig:example1}(a) prevents us from determining the direction of arrows. To this end, we propose to explore the intrinsic causal mechanism behind the spurious correlation as follows:
\begin{proposition}
    \label{proposition 1} \textbf{(Latent common cause~\cite{Verma1993})}
    In causal model $M$, the spurious correlation between variables $S$ and $Y$: $S - Y$ represents a latent common cause $S\leftarrow L\rightarrow Y$.
\end{proposition}
Proposition \ref{proposition 1} leads us to the DAG depicted in Figure \ref{fig:example1}(b), eliminating $S - Y$, thus enabling the analysis of backdoor paths in the causal graph. For ease of understanding, we construct a biased ColoredMNIST example to demonstrate the fork-specific spurious correlation caused by latent common cause. $L$ represents a domain-specific rule that specifies the combination of a particular color and a specific label. Under this setting, we observe that the correlation between $S$ and $Y$ varies with different values of $L=l$. $L=0,1,2,3$, respectively, yielding combinations $(Y=0,S=Red),(Y=0,S=Green),(Y=1,S=Red),(Y=1,S=Green)$. However, the dataset $X$ is biased, resulting in a scenario where 90$\%$ of digit 0 is red and 10$\%$ is green, while 90$\%$ of digit 1 is green and 10$\%$ is red. In this case, the bias degree is 0.9.

\paragraph{Theoretical Analysis of Fork-Specific Spurious Correlation.}
To determine the correctness of the fork-specific spurious correlation in OOD generalization, we impose the following causal analysis.

\begin{definition}
    \label{definition 1} \textbf{(Non-confounding~\cite{stone1993assumptions,robins1997causal})}
    Let $T$ be the set of all variables unaffected by $X$, composed of two disjoint subsets $T_1$ and $T_2$. $X$ and $Y$ are non-confounding if and only if:
    \begin{enumerate}
        \item $T_1$ is independent of $X$: $\mathbb{P}(X)=\mathbb{P}(X\mid T_1)$
        \item Given $X$ and $T_1$, $T_2$ is independent of $Y$: $\mathbb{P}(Y\mid T_1,X)=\mathbb{P}(Y\mid T_1,T_2,X)$
    \end{enumerate}
\end{definition}
By Definition \ref{definition 1}, we know that in Figure \ref{fig:example1}(b), when $T_1=L$ and $T_2=\left \{ X,S \right \}$, $C$ is independent of $L$, and given $\left \{ C,L \right \}$, $\left \{ X,S \right \}$ is independent of $Y$. Therefore, we can conclude that $C\rightarrow Y$ is non-confounding in the fork-specific spurious correlation. Details of the proof are provided in Appendix \ref{B.1}. 
\begin{theorem}
        \label{theorem 1} \textbf{(Non-confounding can achieve OOD generalization)}
    For each domain $D^{v} \in \mathcal{D}$, the data $(X^v, Y^v)$ follows the distribution $\mathbb{P}^{v}\left ( X,Y \right )$. If $C\rightarrow Y^v$ satisfies the assumption of non-confounding, then the invariant features $C=\Phi ^{*}(X^v)$, learned from $X^v$, satisfy the property that $\mathbb{E}(Y^v\mid \Phi ^{*}(X^v))$ is consistent across all domains $D^{v} \in \mathcal{D}$.
\end{theorem}
An intuitive explanation is that in identifying $C\rightarrow Y$, without interference from confounder, it is akin to a randomized controlled experiment. All factors affecting the variation in $Y$, except for $C$, are randomly changing, satisfying the definition of exogenous variables~\cite{pearl2016causal}. Therefore, the invariant features $C$, learned from dataset $X$, are not influenced by the spurious features $S$ across different domains. See Appendix \ref{B.2} for detailed proof.

\begin{wrapfigure}{r}{0.6\textwidth}
\vspace{-0.7cm}
\centering
\includegraphics[width=0.6\textwidth]{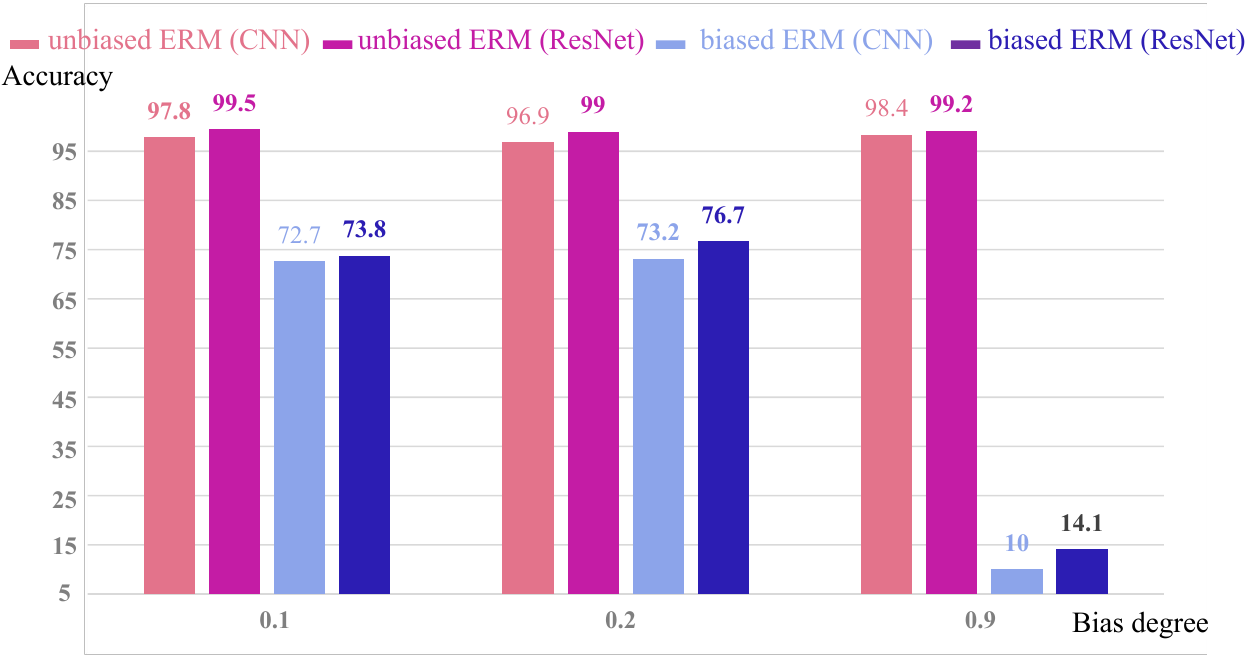}
\vspace{-0.8cm}
\caption{On ColoredMNIST dataset, when the distribution of the training and test sets is consistent (i.e., no bias), ERM achieves an accuracy rate of over 95\%, even the labels are associated with colors to varying degrees. However, when the dataset is biased, associating labels with colors leads to a sharp decline in the accuracy of ERM.}
\label{fig:scm}
\vspace{-0.3cm}
\end{wrapfigure}
However, experiments conducted on biased ColoredMNIST reveal that when the bias level of the training set is 0.9, and the test set remains unbiased, the model fails to address the OOD generalization problem. Detailed experimental results are shown in Figure \ref{fig:scm}. This contradicts the property of non-confounding in the fork-specific spurious correlation (Figure \ref{fig:example1}(b)). 
\textit{This phenomenon prompts us to question whether the proposition that spurious correlation is caused by latent common cause in the field of causal inference is erroneous. }
In the next section, we propose a new model for spurious correlation: \textbf{collider-specific spurious correlation}, to model the general OOD generalization.

\begin{figure}
    \centering
    \includegraphics[width=0.9\textwidth]{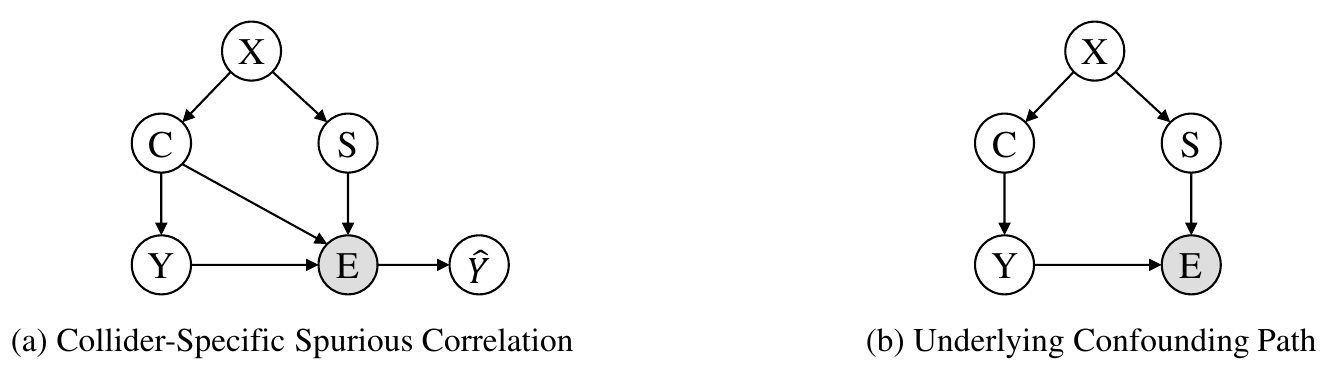}
    \caption{The SCM graph of collider-specific spurious correlation model, with grey nodes indicating conditioning on that variable. Figure (b) is the extracted confounding path that introduces bias in estimating $C \rightarrow Y$.}
    \label{fig:example2}
\end{figure}

\subsection{Causal graph with spurious correlation on OOD Generalization}

\paragraph{Collider-Specific Spurious Correlation.}
Considering a collider structure $Y\rightarrow E\leftarrow S$, we know that $Y \upmodels S$. However, $Y$ and $S$ may be dependent on each other given $E$: for some $y,s,e$, $\mathbb{P}(Y=y\mid S=s,E=e)\neq \mathbb{P}\left ( Y=y\mid E=e \right )$. Therefore, the spurious correlation $S - Y$ can be represented not only by the latent common cause $Y\leftarrow L\rightarrow S$, but also by the conditional common effect\footnote{\underline{E} represents conditioning on E, i.e., E is given.} $Y\rightarrow \underline{E}\leftarrow S$.

As shown in Figure \ref{fig:example2}(a), variables $X$, $C$, $S$, and $Y$ are defined as before, where $E$ represents the learned embedding, and $\hat{Y}$ denotes the predicted result. The main difference compared to the SCM graph in Figure \ref{fig:example1}(a) is that $E=f_E(Y,C,S,\epsilon _{E})$, and $E$ is given. The structural equation for $E$ can be interpreted as learning the representation $E$ in supervised learning using the true label $Y$ and features $C$ and $S$ obtained from the dataset $X$. Since we only need to analyze the confounding between $C$ and $Y$, we extract the confounding path from Figure \ref{fig:example2}(a) to obtain Figure \ref{fig:example2}(b). 

\paragraph{Theoretical Analysis of Collider-Specific Spurious Correlation.}
We first analyze the confounding of $C\rightarrow Y$ in collider-specific spurious correlation. Consider its backdoor path: $C\leftarrow X\rightarrow S\rightarrow \underline{E}\leftarrow Y$. Since $E$ is a collider, conditioning on $E$ opens the backdoor path from $C$ to $Y$, indicating that $C\rightarrow Y$ is confounded. According to Theorem \ref{theorem 1}, the model cannot address the OOD generalization problem. This is also consistent with the empirical results of our experiments on biased ColoredMNIST.

To correctly identify the causal effect of $C\rightarrow Y$, we need to perform the adjustment on confounder using the backdoor criterion~\cite{pearl1993bayesian}. Specifically, conditioning on $S$ can block the backdoor path from $C$ to $Y$, and thus we perform the adjustment on the spurious feature $S$. It is worth emphasizing the theoretical contribution of determining the 
collider-specific spurious correlation, since \textit{adjusting incorrect covariates may lead to biased results}~\cite{grayson1987confounding, kyriacou2016confounding,weinberg1993toward}.
For instance, in the model depicted in Figure \ref{fig:example1}(b) under the assumption of a latent common cause, if we perform the adjustment on the covariate $S$, conditioning on $S$ will open the path $C\leftarrow X\rightarrow \underline{S}\leftarrow L\rightarrow Y$. This path is not a causal path from $C$ to $Y$, since it is a spurious path. In Appendix \ref{C}, we provide a detailed explanation of the issues that arise from incorrectly adjusting covariates. Therefore, we need to carefully consider the correct modeling for the correlation between variables. Accordingly, we explore to perform the adjustment on $S$ to control confounding bias in OOD generalization problems.

\section{Controlling Confounding Bias on OOD Generalization}
\label{others}
In this section, we firstly define the empirical risk minimization (ERM) model based on propensity scores for the OOD generalization problem. Secondly, we employ Fast Fourier Transform (FFT)~\cite{nussbaumer1982fast} and K-means clustering to estimate the propensity scores. We apply the pairing scheme to expand the available samples, thereby obtaining a more accurate estimation of the propensity scores. Finally, our proposed method can be integrated into any existing OOD generalization model.
\subsection{Propensity-Scored ERM for OOD Generalization}
Based on the theoretical analysis above, we identified the spurious feature $S$ as the confounder in $C\rightarrow Y$. We employ backdoor adjustment~\cite{pearl1993bayesian} to eliminate the influence of the confounder $S$ in order to maximize the probability of the invariant feature $C=c$ with respect to the label $Y=y$:
\begin{equation}
    \mathbb{P}\left ( Y=y \mid do(C=c\right ))=\sum _{S=s}\mathbb{P}(Y=y\mid C=c,S=s)\mathbb{P}(S=s).
\end{equation}
Multiply the expression inside the summation by the propensity score~\cite{rosenbaum1983central} $\mathbb{P}(C=c\mid S=s)$, and then divide by this score to obtain:
\begin{equation}
    \mathbb{P}\left ( Y=y \mid do(C=c\right ))=\sum _{S=s}\frac{\mathbb{P}(Y=y\mid C=c,S=s)\mathbb{P}(S=s)\mathbb{P}(C=c\mid S=s)}{\mathbb{P}(C=c\mid S=s)}.
\end{equation}
In fact, the numerator represents the distribution of $(Y,C,S)$ before intervention, and the equation can be written as:
\begin{equation}
    \mathbb{P}\left ( Y=y \mid do(C=c\right )=\sum _{S=s}\frac{\mathbb{P}(Y=y,C=c,S=s)}{\mathbb{P}(C=c\mid S=s)}.
\end{equation}
The probability of each $(Y=y, C=c, S=s)$ in the overall data is amplified by the factor $1/\mathbb{P}(C=c\mid S=s)$. This motivates us to derive a straightforward method to estimate $\mathbb{P}\left ( Y=y \mid do(C=c\right )$ when using finite samples. By weighting each available sample with the factor $1/\mathbb{P}(C=c\mid S=s)$, we achieve estimation of causal effects.
Following the derivation above, we propose the Propensity Score Weighted (PSW) estimator.
\begin{definition}
    \label{definition 2} \textbf{(PSW estimator)}
    The Propensity Score Weighted (PSW) estimator utilizes the propensity scores $\pi(\mathbf{x_{i}^{v}})= \mathbb{P}(C=c_i\mid S=s_i)$ corresponding to the samples $\mathbf{x_{i}^{v}}$ in $D^{v}$ to estimate the expected $\ell$-risk of the invariant function $f$ in a specific domain:
    \begin{equation}
        \mathcal{R}^{PSW}(f|\pi)=\mathbb{E}_{(\mathbf{x},y)\in D^{v}}[\frac{\ell(f(\mathbf{x}),y)}{\pi \left ( \mathbf{x} \right )}].
    \end{equation}
\end{definition}

\begin{definition}
    \label{definition 3} \textbf{(Propensity-scored ERM for OOD generalization)}
    Let $f \in \mathcal{H}$, where $\mathcal{H}\subset \left \{ f:X\rightarrow \mathbb{N} \right \}$ is the hypothesis space of the invariant function $f$. Let $\ell:Y^{v}\times \mathbb{N}\rightarrow \mathbb{R}_{+}$ be the loss function. Then, given a specific domain dataset $D^{v}$, we compute its empirical risk:
    \begin{equation}
\hat{\mathcal{R}}^{PSW}(f|\pi)=\frac{1}{N_v}\sum _{(\mathbf{x_i},y_i)\in D^{v}}[\frac{\ell(f(\mathbf{x_i}),y_i)}{\pi \left ( \mathbf{x_i} \right )}].
    \end{equation}
\end{definition}
Therefore, given the propensity score $\pi(\mathbf{x_{i}})$, we can learn a cross-domain invariant predictor from domain-specific data by minimizing the empirical risk:
\begin{equation}
    \hat{f}\leftarrow \mathop{\arg \min}_{f\in \mathcal{H}}\hat{\mathcal{R}}^{PSW}(f|\pi).
\end{equation}

\subsection{Calculation of Propensity Scores}

\paragraph{Separate Invariant Features and Spurious Features via FFT.}
The literature~\cite{huangfrequency} reports that domain-specific features are mainly contained in the extremely high-frequency and low-frequency signals of images, with the high-frequency signals containing more semantic information. In the domain generalization literature, many FFT-based techniques~\cite{lin2023deep,xu2021fourier} have been proposed to separate causal components from domain-specific information. We posit that the invariant feature $C$ can be represented by the high-frequency components of images, while the spurious feature $S$ is represented by their low-frequency components.

We adopt FFT to extract the spectrum $\mathscr{F}(\mathbf{x_i})$ of the input image $\mathbf{x_i}\in \mathbb{R}^{H\times W}$. Subsequently, we separate the low-frequency part $\mathscr{F}^{l}(\mathbf{x_i})$ and high-frequency part $\mathscr{F}^{h}(\mathbf{x_i})$ in the frequency domain using two binary mask matrices $\mathbf{m}\in \left \{ 0,1 \right \}^{H\times W}$, resulting in corresponding high-pass filtering $\mathcal{M}_{h}^{S}$ and low-pass filtering $\mathcal{M}_{l}^{S}$, where the filter size is denoted by $S$:
\begin{equation}
    \left.\mathcal{M}_l^S(\mathscr{F}(\mathbf{x}))=\boldsymbol{m}\odot\mathscr{F}(\mathbf{x}),\text{ where }\boldsymbol{m}_{i,j}=\left\{\begin{array}{l}1,\text{ if }\min(|i-\frac{H}{2}|,|j-\frac{W}{2}|)\leqslant\frac{S}{2}\\0,\text{ otherwise}\end{array}\right.\right.
\end{equation}

\begin{equation}
    \left.\mathcal{M}_h^S(\mathscr{F}(\mathbf{x}))=\boldsymbol{m}\odot\mathscr{F}(\mathbf{x}),\text{ where }\boldsymbol{m}_{i,j}=\left\{\begin{array}{l}0,\text{ if }\min(|i-\frac{H}{2}|,|j-\frac{W}{2}|)\leqslant\frac{\min(H,W)-S}{2}\\1,\text{ otherwise}\end{array}\right.\right..
\end{equation}
Here, $\odot$ denotes element-wise multiplication, and $\boldsymbol{m}_{i,j}$ denotes the value of $\mathbf{m}$ at position $(i,j)$. Then, we have $\mathscr{F}^{l}(\mathbf{x_i})=\mathcal{M}_l^S(\mathscr{F}(\mathbf{x}))$ and $\mathscr{F}^{h}(\mathbf{x_i})=\mathcal{M}_h^S(\mathscr{F}(\mathbf{x}))$.
We perform inverse Fourier transforms on $\mathscr{F}^{l}(\mathbf{x_i})$ and $\mathscr{F}^{h}(\mathbf{x_i})$ separately to obtain the low-frequency image $\mathbf{x_i^l}=\mathscr{F}^{-1}\circ \mathscr{F}^{l}(\mathbf{x_i})$ and the high-frequency image $\mathbf{x_i^h}=\mathscr{F}^{-1}\circ \mathscr{F}^{h}(\mathbf{x_i})$. Then, we use the encoder $g(\cdot )$ of the invariant function $f(\cdot )$ to extract invariant features and spurious features. Therefore, we obtain the invariant feature $c_i=g(\mathbf{x_i^h})$ and the spurious feature $s_i=g(\mathbf{x_i^l})$ of the image $\mathbf{x_i}$, denoted as $\mathbf{x_i}\sim (c_i,s_i)$.

The propensity score $\mathbb{P}(C=c_i\mid S=s_i)$ for an individual sample point $\mathbf{x_i}$ is difficult to compute. Hence, we first consider a mechanism for designing propensity scores for the population. Initially, we employ the K-means algorithm~\cite{lloyd1982least} to cluster the feature sets $\left \{ c_i \right \}_{i=1}^{N_v}$ and $\left \{ s_i \right \}_{i=1}^{N_v}$. The categories of the invariant feature $C$ are classified based on the number of categories $m$ of the label $Y\in\left \{ 1,2,...,m \right \}$. As for the spurious feature $S$, we use a hyperparameter $n$ to adjust its categories. Let the set of clusters for the invariant feature $C$ be denoted as $\left \{ C^{(1)},C^{(2)},...,C^{(m)} \right \}$, where $c_i \in C^{(k)}$. Similarly, for the spurious feature $S$, the set of clusters after clustering is denoted as $\left \{ S^{(1)},S^{(2)},...,S^{(n)} \right \}$, where $s_i \in S^{(l)}$. Then, the propensity score for the sample point $\mathbf{x_i}$ corresponds to the population is:
\begin{equation}
    \pi \left ( \mathbf{x_i} \right )=\mathbb{P}(C=C^{(k)}\mid S=S^{(l)}).
\end{equation}

\paragraph{Pairing Scheme for Improving Propensity Score Estimation.} To improve the accuracy of estimating the propensity score $\pi \left ( \mathbf{x_i} \right )$, we aim to enhance the encoder $g(\cdot )$'s ability to identify potential spurious features. Therefore, we augment the samples by randomly pairing of $C$ and $S$ to increase the completeness of the available value set of data. Formally, given an original image $\mathbf{x_i}$ and an image $\mathbf{x_j}$ randomly sampled from any source domains, their spectra in the frequency domain are denoted as $\mathscr{F}(\mathbf{x_i})=\mathscr{F}^{l}(\mathbf{x_i})+\mathscr{F}^{h}(\mathbf{x_i})$ and $\mathscr{F}(\mathbf{x_j})=\mathscr{F}^{l}(\mathbf{x_j})+\mathscr{F}^{h}(\mathbf{x_j})$ respectively. We perform linear interpolation on the low-frequency parts of $\mathbf{x_i}$ and $\mathbf{x_j}$, while retaining the high-frequency parts of the original image $\mathbf{x_i}$ to obtain the spectrum of the ``simulated sample'' $\mathbf{\tilde{x}_i}$:
\begin{equation}
    \mathscr{F}(\mathbf{\tilde{x}_i})=\mathscr{F}^{h}(\mathbf{x_i})+(1-\lambda )\mathscr{F}^{l}(\mathbf{x_i})+\lambda \mathscr{F}^{l}(\mathbf{x_j})
\end{equation}
where $\lambda \sim U(0,\delta )$, with $\delta$ controlling the mixing ratio. Applying the inverse Fourier transform to $\mathscr{F}(\mathbf{\tilde{x}_i})$, we obtain the ``simulated sample'' $\mathbf{\tilde{x}_i}=\mathscr{F}^{-1}\circ \mathscr{F}(\mathbf{\tilde{x}_i})$. For each sample $\mathbf{x_i^{v}}$ in the training dataset $D^{v}=\left \{ \left ( \mathbf{x_{i}^{v}},y_{i}^{v} \right ) \right \}_{i=1}^{N_{v}}$, we add its corresponding $\mathbf{\tilde{x}_i^{v}}$ to obtain $\tilde{D}^{v}=\left \{ \left ( \mathbf{x_{i}^{v}},y_{i}^{v} \right ),\left ( \mathbf{\tilde{x}_{i}^{v}},y_{i}^{v} \right ) \right \}_{i=1}^{N_{v}}$. By training the invariant function $f(\cdot)$ on $\tilde{D}^{v}$, we can update the parameters of the encoder $g(\cdot)$, thereby more accurately extracting $c_i$ and $s_i$:
\begin{equation}
    \mathcal{L}^{PS}=\frac{1}{2 N_v}\sum _{(\mathbf{x_i},y_i)\in \tilde{D}^{v}}\ell(f(\mathbf{x_i}),y_i).
\end{equation}

\paragraph{Learning Objectives.} For any existing OOD method, we assume that the learned invariant function is denoted as $f(\cdot)$ and the encoder for the original images $\mathbf{x_i}$ is $g(\cdot)$. Utilizing the PSW estimator, we obtain the loss term $\mathcal{L}^{PSW}(D^{v})$ in the original training domain $D^{v}$. $\mathcal{L}^{PS}(\tilde{D}^{v})$ is trained on the augmented samples $\tilde{D}^{v}$ for a more precise estimation of the propensity score $\pi \left ( \mathbf{x} \right )$. $\mathcal{L}(D^{v})$ preserves the training loss of the original method. The final learning objective is defined as: 
\begin{equation}
    \mathcal{L}_{our}=\mathcal{L}(D^{v})+\alpha \mathcal{L}^{PSW}(D^{v})+\beta \mathcal{L}^{PS}(\tilde{D}^{v}).
\end{equation}

\section{Theoretical Analysis for PSW Estimator}
We analyze the Error Bound of the PSW estimator on the OOD generalization through Corollary \ref{corollary 1}, quantifying the gap between the true risk $\mathcal{R}(\hat{f})$ and the empirical risk $\hat{\mathcal{R}}^{PSW}(\hat f|\pi)$ caused by the propensity score $\pi(\mathbf{x})$. The detailed proof is provided in Appendix \ref{D}, following the approach outlined in~\cite{cortes2010learning}.

\begin{corollary}
    \label{corollary 1} \textbf{(Propensity-scored ERM for OOD generalization error bound)}
    For any finite hypothesis space $\mathcal{H}=\left \{ \hat{f_1},...,\hat{f}_{\left | \mathcal{H} \right |} \right \}$ of invariant functions $\hat{f}$, and a loss $0\leq \ell(\hat{f}(\mathbf{x}),y)\leq \Omega $, the true risk $\mathcal{R}(\hat{f})$ of the $\hat{f}$ that minimizes empirical risk in hypothesis space $\mathcal{H}$, has an upper bound with probability $1-\delta$ on a specific domain $D^{v}$:
    \begin{equation}
        \mathcal{R}(\hat{f})\leq \hat{\mathcal{R}}^{PSW}(\hat f|\pi)+\frac{\Omega }{N_v}\sqrt{\frac{\log(2\left | \mathcal{H} \right |/\delta )}{2}}\sqrt{\sum_{i=1}^{N_v}\frac{1}{\pi^2(\mathbf{x_i})}}.
    \end{equation}
\end{corollary}

\begin{table*}
\caption{The domain generalization accuracy beyond the training domain are evaluated across six distinct datasets. Results are presented as the average accuracy across all test environments, with the standard deviation determined over three independent runs. The validation approach employed adheres to the training domain validation protocol commonly utilized in OOD field.}
\vspace{-0.4cm}
\begin{center}
\adjustbox{max width=\textwidth}{%
\begin{tabular}{lccccccc}
\toprule
\textbf{Algorithm}        & \textbf{ColoredMNIST}     & \textbf{RotatedMNIST}     & \textbf{VLCS}             & \textbf{PACS}             & \textbf{OfficeHome}       & \textbf{TerraIncognita}  & \textbf{Avg}              \\
\midrule
ERM \cite{vapnik1998statistical}                      
& 51.5 $\pm$ 0.1            & 98.0 $\pm$ 0.0            & 77.5 $\pm$ 0.4       & 85.5 $\pm$ 0.2            & 66.5 $\pm$ 0.3            & 46.1 $\pm$ 1.8       & 70.9 \\
IRM  \cite{arjovsky2019invariant}                   
& 52.0 $\pm$ 0.1            & 97.7 $\pm$ 0.1            & 78.5 $\pm$ 0.5       & 83.5 $\pm$ 0.8            & 64.3 $\pm$ 2.2            & 47.6 $\pm$ 0.8       & 70.6 \\
GroupDRO  \cite{sagawa2019distributionally}
& 52.1 $\pm$ 0.0            & 98.0 $\pm$ 0.0            & 76.7 $\pm$ 0.6       & 84.4 $\pm$ 0.8            & 66.0 $\pm$ 0.7            & 43.2 $\pm$ 1.1       & 70.1 \\
Mixup \cite{yan2020improve}                  
& 52.1 $\pm$ 0.2            & 98.0 $\pm$ 0.1            & 77.4 $\pm$ 0.6       & 84.6 $\pm$ 0.6            & 68.1 $\pm$ 0.3            & 47.9 $\pm$ 0.8       & 71.4 \\
MLDG    \cite{MLDG}                  & 51.5 $\pm$ 0.1            & 97.9 $\pm$ 0.0            & 77.2 $\pm$ 0.4            & 84.9 $\pm$ 1.0            & 66.8 $\pm$ 0.6            & 47.7 $\pm$ 0.9                        & 71.0                      \\
CORAL    \cite{CORAL}       & 51.5 $\pm$ 0.1            & 98.0 $\pm$ 0.1            & \textbf{78.8} $\pm$ 0.6            & 86.2 $\pm$ 0.3            & 68.7 $\pm$ 0.3            & 47.6 $\pm$ 1.0                        & 71.8                      \\
MMD    \cite{MMD}     & 51.5 $\pm$ 0.2            & 97.9 $\pm$ 0.0            & 77.5 $\pm$ 0.9            & 84.6 $\pm$ 0.5            & 66.3 $\pm$ 0.1            & 42.2 $\pm$ 1.6                       & 70.0                      \\
DANN   \cite{DANN}                   & 51.5 $\pm$ 0.3            & 97.8 $\pm$ 0.1            & 78.6 $\pm$ 0.4            & 83.6 $\pm$ 0.4            & 65.9 $\pm$ 0.6            & 46.7 $\pm$ 0.5                      & 70.7                     \\
CDANN             \cite{CDANN}        & 51.7 $\pm$ 0.1            & 97.9 $\pm$ 0.1            & 77.5 $\pm$ 0.1            & 82.6 $\pm$ 0.9            & 65.8 $\pm$ 1.3            & 45.8 $\pm$ 1.6                        & 70.2                      \\
MTL                  \cite{MTL}     & 51.4 $\pm$ 0.1            & 97.9 $\pm$ 0.0            & 77.2 $\pm$ 0.4            & 84.6 $\pm$ 0.5            & 66.4 $\pm$ 0.5            & 45.6 $\pm$ 1.2                        & 70.5                      \\
SagNet     \cite{SagNet}               & 51.7 $\pm$ 0.0            & 98.0 $\pm$ 0.0            & 77.8 $\pm$ 0.5            & 86.3 $\pm$ 0.2            & 68.1 $\pm$ 0.1            & \textbf{48.6} $\pm$ 1.0                       & 71.8                      \\
ARM                     \cite{arm}  & 56.2 $\pm$ 0.2            & \textbf{98.2} $\pm$ 0.1            & 77.6 $\pm$ 0.3            & 85.1 $\pm$ 0.4            & 64.8 $\pm$ 0.3            & 45.5 $\pm$ 0.3                        & 71.2                      \\
VREx      \cite{vrex} & 51.8 $\pm$ 0.1            & 97.9 $\pm$ 0.1            & 78.3 $\pm$ 0.2            & 84.9 $\pm$ 0.6            & 66.4 $\pm$ 0.6            & 46.4 $\pm$ 0.6                      &      71.0                 \\
RSC            \cite{rsc}           & 51.7 $\pm$ 0.2            & 97.6 $\pm$ 0.1            & 77.1 $\pm$ 0.5            & 85.2 $\pm$ 0.9            & 65.5 $\pm$ 0.9            & 46.6 $\pm$ 1.0                        & 70.6                      \\

Fish \cite{fish} & 51.6 $\pm$ 0.1 & 98.0 $\pm$ 0.0 & 77.8 $\pm$ 0.3                      & 85.5 $\pm$ 0.3 & 68.6 $\pm$ 0.4 & 45.1 $\pm$ 1.3                     & 71.1 \\

Fishr \cite{rame2022fishr} & 52.0 $\pm$ 0.2 & 97.8 $\pm$ 0.0 & 77.8 $\pm$ 0.1                      & 85.5 $\pm$ 0.4 & 67.8 $\pm$ 0.1 & 47.4 $\pm$ 1.6                     & 71.4 \\

AND-mask \cite{and-mask} & 51.3 $\pm$ 0.2 & 97.6 $\pm$ 0.1 & 78.1 $\pm$ 0.9                      & 84.4 $\pm$ 0.9 & 65.6 $\pm$ 0.4 & 44.6 $\pm$ 0.3                     & 70.3 \\

SAND-mask \cite{sand-mask} & 51.8 $\pm$ 0.2 & 97.4 $\pm$ 0.1 & 77.4 $\pm$ 0.2                      & 84.6 $\pm$ 0.9 & 65.8 $\pm$ 0.4 & 42.9 $\pm$ 1.7                      & 70.0 \\

SelfReg \cite{kim2021selfreg} & 52.1 $\pm$ 0.2 & 98.0 $\pm$ 0.1 & 77.8 $\pm$ 0.9                      & 85.6 $\pm$ 0.4 & 67.9 $\pm$ 0.7 & 47.0 $\pm$ 0.3                      & 71.4 \\

$\text{CausIRL}_{\text{CORAL}}$ \cite{caus} & 51.7 $\pm$ 0.1 & 97.9 $\pm$ 0.1 & 77.5 $\pm$ 0.6 & 85.8 $\pm$ 0.1 & 68.6 $\pm$ 0.3 & 47.3 $\pm$ 0.8                      & 71.5 \\

$\text{CausIRL}_{\text{MMD}}$ \cite{caus} & 51.6 $\pm$ 0.1 & 97.9 $\pm$ 0.0 & 77.6 $\pm$ 0.4                      & 84.0 $\pm$ 0.8 & 65.7 $\pm$ 0.6 & 46.3 $\pm$ 0.9                      & 70.5 \\

BalancingERM \cite{wang2022causal} & 60.1 $\pm$ 1.0 & 97.7 $\pm$ 0.0 & 76.1 $\pm$ 0.3                      & 85.2 $\pm$ 0.4 & 67.1 $\pm$ 0.4 & 48.0 $\pm$ 1.7                      & 72.3 \\

BalancingCORAL \cite{wang2022causal} & 66.6 $\pm$ 1.2 & 97.7 $\pm$ 0.1 & 76.4 $\pm$ 0.5                      & 86.7 $\pm$ 0.1 & 69.6 $\pm$ 0.2 & 47.0 $\pm$ 1.2                      & 74.0 \\
\midrule
BalancingERM+Ours & 62.5 $\pm$ 2.5 & 98.1 $\pm$ 0.2 & 77.1 $\pm$ 2.2                      & 86.3 $\pm$ 1.2 & 68.2 $\pm$ 0.8 & 48.2 $\pm$ 0.6                 & 73.4 \\
BalancingCORAL+Ours & \textbf{67.8} $\pm$ 1.6 & 97.8 $\pm$ 0.1 & 77.3 $\pm$ 1.8                      & \textbf{87.4} $\pm$ 0.4 & \textbf{70.1} $\pm$ 1.2 & 48.3 $\pm$ 1.3  & \textbf{74.8} \\
\bottomrule
\end{tabular}}
\end{center}
\label{tab:main_results}
\end{table*}

\section{Experiments}
\noindent \textbf{Datasets and baselines.} To evaluate the performance of our proposed method, we select six canonical domain generalization benchmarks, which include \textbf{ColoredMNIST} \cite{ahuja2020empirical}, \textbf{RotatedMNIST} \cite{rmnist}, \textbf{VLCS} \cite{VLCS}, \textbf{PACS} \cite{pacs}, \textbf{OfficeHome} \cite{officehome}, and \textbf{TerraIncognita} \cite{beery2018recognition}. For a comprehensive comparison, 21 benchmark domain generalization methods are chosen as our baselines. It is noteworthy that Balancing \cite{wang2022causal} is the prior state-of-the-art method, which proposes a balanced mini-batch sampling method.
To regulate the influence exerted by the foundational algorithms, a uniform set of hyper-parameters is employed across both the baselines and our proposed methods. We utilize train domain validation for model selection due to its practical applicability as a validation technique. A comprehensive exposition regarding the datasets is provided in the \textbf{Appendix} \ref{sec:experimental_details}.

\begin{figure}[t]
\vspace{-0.5cm}
    \centering
    \includegraphics[width=1\textwidth]{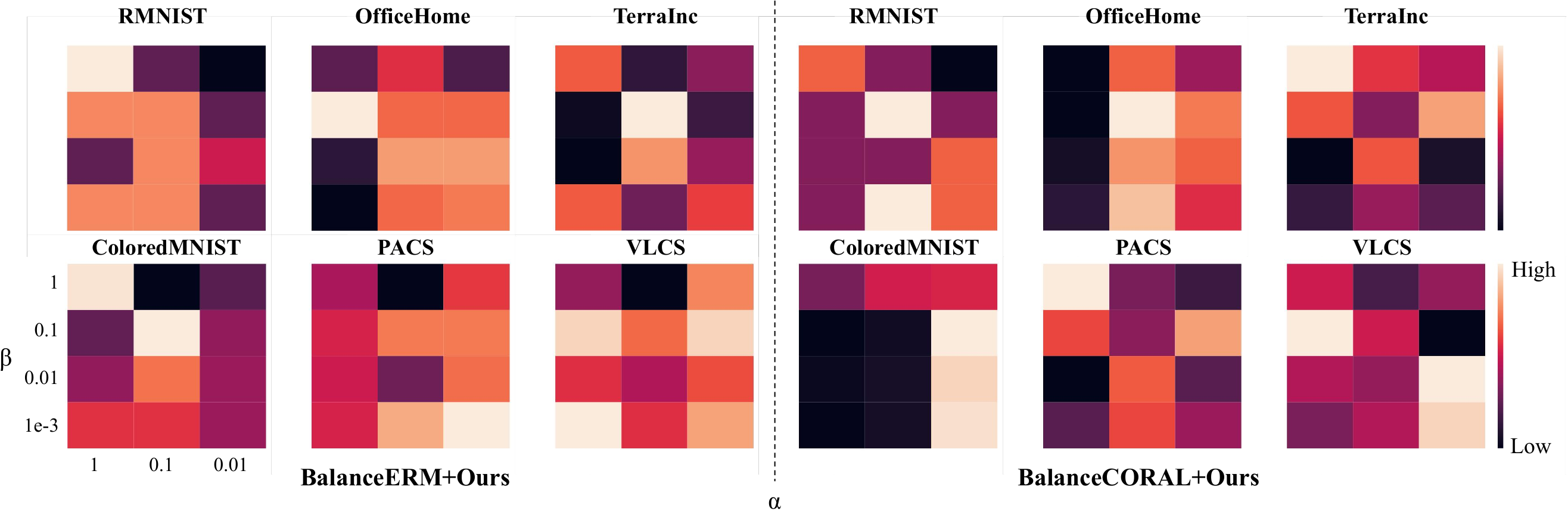}
    \vspace{-0.5cm}
    \caption{The results of hyper-parameters.}    
    \label{fig:heatmap}
    \vspace{-0.6cm}
\end{figure}

\noindent \textbf{Domian generalization results.} Functioning as a plug-and-play module, our method is applied to two representative
base algorithms: BalancingERM and BalancingCORAL. 
BalancingERM and BalancingCORAL change the sampling strategy of ERM and CORAL without altering the loss function of base models.
As shown in Table \ref{tab:main_results}, our method can significantly improve the domain generalization performance of benchmark methods, e.g., our method improves the average performance of BalancingERM by 1.1\% and the average performance of CORAL by 0.8\%. Furthermore, we achieve the sota domain generalization performance on ColoredMNIST, PACS, and OfficeHome. The observations demonstrate the superiority of proposed method. Meanwhile, the results showcase that the adjustment of variable $S$ can improve the performance of domain generalization, which contradicts the fork-specific spurious correlation in Figure \ref{fig:example1}
(b), thereby further verifying the correctness of collider-specific spurious correlation in Figure \ref{fig:example2}(b).

\noindent \textbf{Ablation study.} We propose two loss functions for domain generalization, i.e., $\mathcal{L}_{PSW}$ and $\mathcal{L}_{PS}$. We conduct ablation study to demonstrate the effectiveness of each proposed technique. Concretely, we trained two variants for both BalancingERM+Ours and BalancingCORAL+Ours, i.e., BalancingERM+Ours w/o $\mathcal{L}_{PSW}$, BalancingERM+Ours w/o $\mathcal{L}_{PS}$, BalancingCORAL+Ours w/o $\mathcal{L}_{PSW}$, and BalancingCORAL+Ours w/o $\mathcal{L}_{PS}$. The results are depicted in Table \ref{tab:ablation study}. We can observe that regardless of which loss function is removed, the performance of the model decreases, thereby demonstrating the effectiveness of each proposed technique. Furthermore, BalancingERM+Ours w/o $\mathcal{L}_{PS}$ (BalancingCORAL+Ours w/o $\mathcal{L}_{PS}$) typically outperforms BalancingERM+Ours w/o $\mathcal{L}_{PSW}$ (BalancingCORAL+Ours w/o $\mathcal{L}_{PSW}$), indicating that debiasing methods based on propensity scores are benifical for enhancing the model's generalizability.
 
\noindent \textbf{Hyper-parameter research.} There are two hyper-parameters in our method, i.e., $\alpha$ and $\beta$. $\alpha, \beta$ are searched in $\{1,0.1,0.01\}$ and $\{1,0.1,0.01,0.001\}$, respectively. We determine their specific values through experimental results. In Figure \ref{fig:heatmap}, we depict the results on ColoredMNIST, PACS, and VLCS. The lighter the color, the higher the classification performance. As shown in Figure \ref{fig:heatmap}, the best combination of $\alpha$ and $\beta$ varies with the baselines and datasets, e.g., BalanceERM+Ours achieves its best performance on ColoredMNIST with $\{\alpha=0.1,\beta=0.1\}$ and on PACS with $\{\alpha=0.01,\beta=0.001\}$. Therefore, a detailed assignment of hyper-parameters can help to improve the performance of our model.

\begin{figure}[t]
    \centering
    \includegraphics[width=1\textwidth]{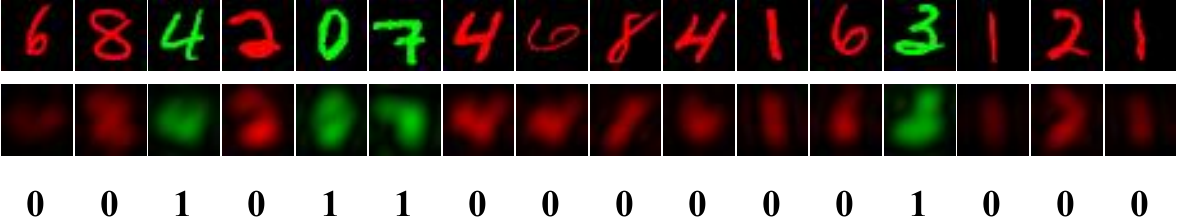}
    \caption{The results of case study on a batch of randomly selected samples.}
    \vspace{-0.4cm}
    \label{fig:case_study}
\end{figure}

\noindent \textbf{Case study.}
Classifying the background semantics is a critical step in computing the loss $\mathcal{L}_{PSW}$, and classification accuracy determines the estimation precision of the propensity scores. We conduct a case study on a batch of randomly selected samples from the ColoredMNIST dataset to assess the precision of calculating propensity scores. As illustrated in Figure \ref{fig:case_study}, our method not only delineates the background semantics clearly but also classifies them accurately. In the ColoredMNIST dataset, where backgrounds are either red or green, therefore the predicted classification labels are 0 and 1. The results of case study demonstrate that our method can accurately estimate the propensity scores. 
\begin{table}
    \centering
        \caption{The ablation study on five datasets.}
        \adjustbox{max width=\textwidth}{
    \begin{tabular}{cccccc}
    \toprule[1.5pt]
        Algorithm & ColoredMNIST & VLCS & PACS & OfficeHome & TerraIncognita \\
        \midrule
BalancingERM w/o $\mathcal{L}_{PS}$  & 61.8 & 75.7 & 85.4 & 67.7 & 46.2 \\
BalancingERM w/o $\mathcal{L}_{PSW}$ & 60.5 & 75.6 & 84.6 & 67.1 & 47.1 \\
\midrule
BalancingERM & 62.5 & 77.1 & 86.3 & 68.2 & 48.2 \\
\midrule[1pt]
BalancingCORAL w/o $\mathcal{L}_{PS}$ & 67.4    &   76.4    &     86.8   &       69.7   &       46.7  \\
BalancingCORAL w/o $\mathcal{L}_{PSW}$ & 67 & 76.1    &  86.3       &   86.4  &    47.7  \\
\midrule
BalancingCORAL & 67.8 & 77.3 & 87.4 & 70.1 & 48.3 \\
    \bottomrule[1.5pt]
    \end{tabular}}
     \vspace{-0.4cm}
    \label{tab:ablation study}
\end{table}

\section{Related Work}
\label{related}
Causal inference-based methods~\cite{pfister2019invariant,cui2020causal,chen2022learning,wu2022discovering,cui2022stable,yang2024invariant} have made significant strides in the field of domain generalization, offering a novel perspective on addressing the out-of-distribution (OOD) problem. These approaches construct appropriate causal models for the data generation process~\cite{arjovsky2019invariant,rosenfeld2020risks,ahuja2021invariance,gamella2020active,oberst2021regularizing,lu2021invariant}, severing the spurious correlations between causal and spurious features to capture stable causal representations invariant across domains or environments, leveraging information from multiple environments. Among these works, the most representative method is Invariant Risk Minimization (IRM), which has strong theoretical guarantees in linear systems. Building upon IRM, numerous variants have emerged~\cite{ahuja2020invariant,krueger2021out,koyama2020invariance,rosenfeld2022online,robey2021model,wang2022provable,zhang2021task}, aiming to address some of IRM's challenges, such as its failure in nonlinear tasks~\cite{rosenfeld2020risks}, the requirement for extensive domain information~\cite{lin2022zin}, and optimization difficulties in deep neural networks~\cite{jin2020domain,chen2022pareto}. Beyond analyzing the fundamental causal mechanisms of OOD data generation, there are also causal methods involving interventions~\cite{mao2021generative,liu2022show,wu2022discovering,mao2022causal}. These methods include robust feature learning through data augmentation by intervening on spurious features~\cite{wu2022discovering} or achieving interventional distributions through frontdoor/backdoor adjustments~\cite{mao2021generative,mao2022causal,liu2022show}.
\section{Conclusions and Limitations}
\label{conclusion}
In this paper, we model the phenomenon of spurious correlations in domain generalization from a novel perspective within the representation learning process. We analyze the underlying causal mechanisms that generate spurious correlations and identify two types: fork-specific and collider-specific spurious correlations. We clarify that the non-confounding between invariant features and true labels is key to addressing OOD problems. Based on this premise, we demonstrate that collider-specific spurious correlation is the correct OOD-oriented SCM. Therefore, motivated by the causal analysis, we control confounding bias by adjusting spurious features, introducing a propensity score weighted estimator to ensure that the causal effect of invariant features on true labels is non-confounding.
However, we cannot entirely eliminate the confounding bias introduced by spurious features, as this requires perfect propensity score calculation. Therefore, achieving accurate propensity score estimation necessitates a bias-variance trade-off, where we aim to avoid excessively small propensity scores to reduce variance while sacrificing some accuracy.

\bibliography{ref}
\bibliographystyle{unsrt}

\appendix
\newpage
\section{Causal Background Knowledge}
\label{A}
\paragraph{Structural Causal Models and Intervention.}
A structural causal model (SCM)~\cite{pearl2009causality} is a triple $M=\left \langle X,U,F \right \rangle$, where $U$ is known as the \textit{exogenous variable}, determined by external factors of the model. $X=\left \{ X_1,X_2,...,X_n \right \}$ is referred to as the \textit{endogenous variable}, whose changes are determined by the functions $F=\left \{ f_1,f_2,...,f_n \right \}$. Each $f_i$ represents $\left \{ f_i:U_i\cup PA_i \rightarrow X_i\right \}$, where $U_i\subseteq U$, $PA_i\subseteq X\backslash X_i$, satisfying:
\begin{equation}
    x_i=f_i\left ( pa_i,u_i \right ),\quad i=1,2,...,n.
\end{equation}
Each causal model $M$ corresponds to a directed acyclic graph (DAG) $G$, where each node corresponds to a variable in $X\cup U$, and directed edges point from $U_i\cup PA_i$ to $X_i$.

An \textit{intervention} refers to forcing a variable $X_i$ to take a fixed value $x_i$. This equivalently removes $X_i$ from the influence of its original functional mechanism $x_i=f_i\left ( pa_i,u_i \right )$ and replaces it with a constant function $X_i=x_i$. Formally, we denote the intervention as $do(X_i=x_i)$, or simply $do(x_i)$. After the intervention on $X_i$, the corresponding causal graph $G_{x_i}$ is obtained by removing all arrows pointing to $X_i$ in $G$ to represent the post-intervention world.

\paragraph{Path and $d$-separation.} We summarize two classic definitions~\cite{pearl2016causal} to help us determine the independence between variables in the SCM graph. They enable us to avoid cumbersome probability calculations and instead obtain independence between variables directly from the graph.

\begin{definition}
    \label{path} \textbf{(Path)}
    In the SCM graph, the paths from variable $X$ to $Y$ include three types of structures: 1) Chain Structure: $A \rightarrow B\rightarrow C$ or $A \leftarrow B\leftarrow C$, 2) Fork Structure: $A \leftarrow B\rightarrow C$, and 3) Collider Structure: $A \rightarrow B\leftarrow C$.
\end{definition}

\begin{definition}
    \label{separation} \textbf{($d$-separation)}
A path $p$ is blocked by a set of nodes $Z$ if and only if:
\begin{enumerate}
    \item $p$ contains a chain of nodes $A \rightarrow B\rightarrow C$ or a fork  $A \leftarrow B\rightarrow C$ such that the middle node $B$ is in $Z$ (i.e., $B$ is conditioned on), or
    \item $p$ contains a collider $A \rightarrow B\leftarrow C$ such that the collider node $B$ is not in $Z$, and no descendant of $B$ is in $Z$. 
\end{enumerate}
\end{definition}
If $Z$ blocks every path between two nodes $X$ and $Y$ , then $X$ and $Y$ are \textit{$d$-separated}, conditional on $Z$, and thus are independent conditional on $Z$, denoted as $X \upmodels Y \mid Z$.

\paragraph{Backdoor and Backdoor Adjustment.} 
\begin{definition}
    \label{Back-Door} \textbf{(Backdoor)}
In a DAG $G$, a set of variables $Z$ satisfies the backdoor criterion for an ordered pair of variables $(X_i, X_j)$ if:
\begin{enumerate}
    \item No node in $Z$ is a descendant of $X_i$.
    \item $Z$ blocks all paths between $X_i$ and $X_j$ that are directed into $X_i$.
\end{enumerate}
Similarly, if $X$ and $Y$ are two disjoint subsets of nodes in $G$, then $Z$ is said to satisfy the backdoor criterion for $(X, Y)$ if $Z$ satisfies the backdoor criterion for any pair of variables $(X_i, X_j)$, where $X_i \in X$ and $X_j \in Y$.
\end{definition}

\begin{definition}
    \label{adjustment} \textbf{(Backdoor adjustment)}
If a set of variables $Z$ satisfies the backdoor criterion for $(X, Y)$, then the causal effect of $X$ on $Y$ is identifiable and can be given by the following formula:
\begin{equation}
    \mathbb{P}(Y=y\mid do(X=x))=\sum _{z}\mathbb{P}(Y=y\mid X=x,Z=z)\mathbb{P}(Z=z).
\end{equation}
\end{definition}

\section{Theoretical Analysis of Causal Models}
\label{B}
\subsection{Fork-Specific Spurious Correlation is Non-Confounding}
\label{B.1}
To investigate whether there is confounding in the relationship $C \rightarrow Y$, we represent the set of variables not influenced by $C$ as $T = \{L, X, S\}$ according to Definition \ref{definition 1}. This can be explained from the perspective of structural equations: variables in $T$ do not change based on the value of $C$, meaning that $T$ does not contain descendant nodes of $C$. We further partition $T$ into two disjoint subsets: $T_1 = \{L\}$ and $T_2 = \{X, S\}$.

First, we verify condition 1 of Definition \ref{definition 1}, which states that $C \upmodels L$. There are two paths from $C$ to $L$: $C \rightarrow Y \leftarrow L$ and $C \leftarrow X \rightarrow S \leftarrow L$. Both are blocked by the collider nodes $Y$ and $S$, respectively, ensuring that $C$ and $L$ are independent.

For condition 2, we need to verify that $\{X, S\} \upmodels Y \mid \{C, L\}$. We begin by showing that $X \upmodels Y \mid \{C, L\}$. The path $X \rightarrow C \rightarrow Y$ is blocked by conditioning on $C$, and the path $X \rightarrow S \leftarrow L \rightarrow Y$ is blocked by the collider $S$. Additionally, conditioning on $L$ does not introduce any new paths. Next, we show that $S \upmodels Y \mid \{C, L\}$. The path $S \leftarrow L \rightarrow Y$ is blocked by conditioning on $L$, and the path $S \leftarrow X \rightarrow C \rightarrow Y$ is blocked by conditioning on $C$.

Since our definition of non-confounding is a mathematically formalized definition from the perspective of statistical associations, we can determine whether the required independencies in the two conditions are satisfied through independence tests, even without using the $d$-separation criterion mentioned above.

\subsection{Non-Confounding Can Achieve OOD Generalization}
\label{B.2}
\begin{figure}
    \centering
    \includegraphics[width=0.9\textwidth]{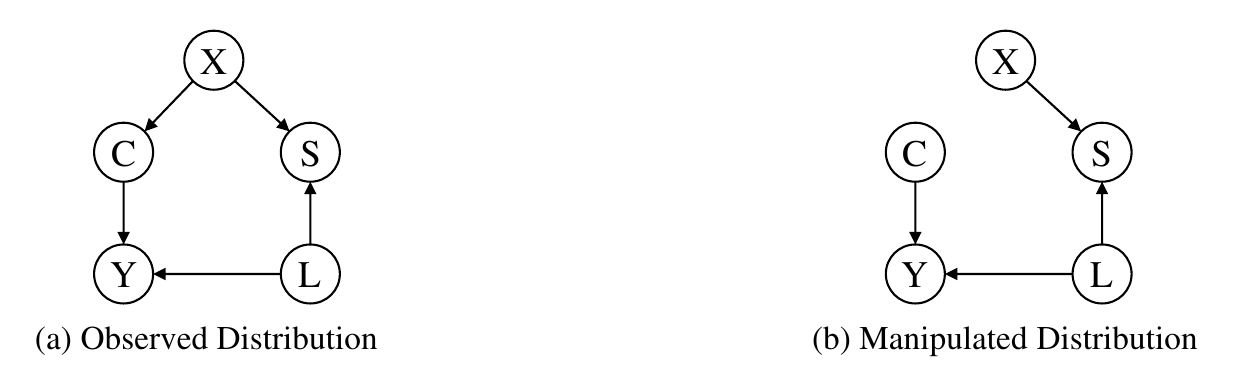}
    \caption{After intervening on variable $C$, remove all arrows pointing to $C$. In the post-intervention world, variables follow the manipulated distribution $\mathbb{P}_m$, meaning that $C$ and $S$ are $d$-separated.}
    \label{fig:example3}
\end{figure}

We provide a mathematically formalized causal definition of non-confounding, which addresses the limitations of using statistical methods to test for confounding.
\begin{definition}
    \label{do} \textbf{(Non-confounding: causal definition)}
    Let $\mathbb{P}(y\mid do(c))$ denote the probability of the response event $Y=y$ under the intervention $C=c$. We say that $C$ and $Y$ are non-confounding if and only if:
    \begin{equation}
        \mathbb{P}(y\mid do(c))=\mathbb{P}(y\mid c).
    \end{equation}
\end{definition}
The meaning of the probability $\mathbb{P}(y\mid do(c))$ can be explained by structural equations. This is equivalent to removing $C$ from its original functional mechanism $C=f_C(pa_C, \epsilon_C)$, and modifying this function to a constant function $C=c$. That is, any variables influencing the value of $C$ become independent of $C$ in the intervened world. This intervened world can be understood as a randomized controlled experiment with respect to $C$, and we denote the probability in this world as the \textit{manipulated probability} $\mathbb{P}_m$.

In different domains $D^{v_1}$ and $D^{v_2}$, domain-specific spurious features $S$ are independent of $C$, i.e., $\mathbb{P}_m(C) = \mathbb{P}_m(C\mid S^{v_1}) = \mathbb{P}_m(C\mid S^{v_2})$. For example, in the case of fork-specific spurious correlation, the intervened world is represented as in Figure \ref{fig:example3}(b), where we can see that $S$ and $C$ are $d$-separated. Thus, $\mathbb{E}_{(C,S)\sim \mathbb{P}_m}(Y^v\mid \Phi^*(X^v))$ is invariant across domains. From the Eq.(17), we have $\mathbb{P}(y\mid do(c))=\mathbb{P}_m(Y\mid C) = \mathbb{P}(Y\mid C)$. Therefore, we can conclude that $\mathbb{E}_{(C,S)\sim \mathbb{P}}(Y^v\mid \Phi^*(X^v))$ is also invariant across domains.

\section{Discussion on Adjusting Erroneous Covariates}
\label{C}
In this section, we discuss the distinction between statistical correlation and causal confounding: the correlation between covariate $Z$ and variables $C$ and $Y$ does not imply confounding between $C$ and $Y$. Incorrect adjustment for covariate $Z$ may result in bias. Therefore, when analyzing the causal effect of $C \rightarrow Y$, it is necessary to select specific models: fork-specific spurious correlation or collider-specific spurious correlation, in order to control for confounding bias.

\begin{figure}
    \centering
     \vspace{-5ex}
    \includegraphics[width=\textwidth]{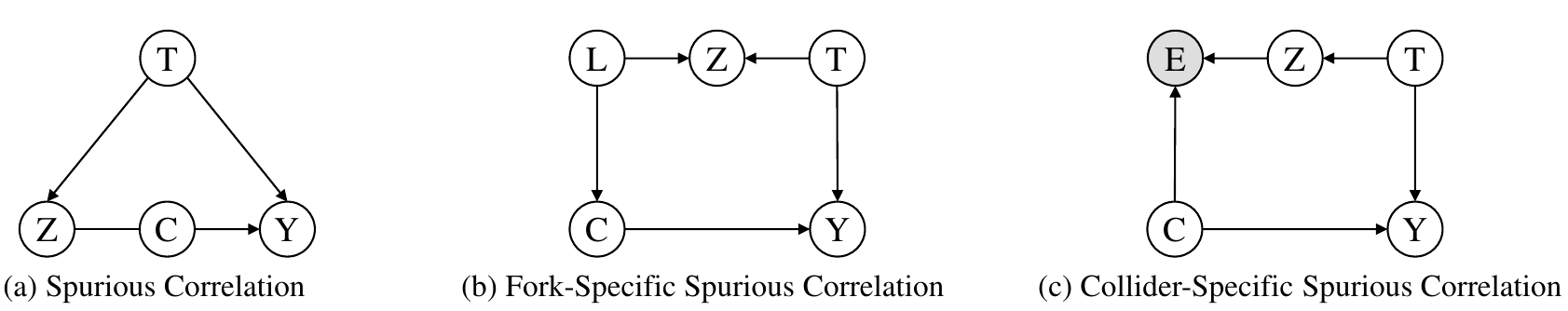}
    \caption{$Z$ and $C$ form a spurious correlation $Z - C$ as shown in Figure (a). Figures (b) and (c) model this correlation according to fork-specific spurious correlation and collider-specific spurious correlation, respectively. Although the variable correlations in (b) and (c) are the same, the confounding between $C$ and $Y$ is completely different. Figure (b) is non-confounding, whereas Figure (c) is confounding.}
     \vspace{-2ex}
    \label{fig:example4}
\end{figure}

\begin{figure}[htbp]
    \centering
    \includegraphics[width=\textwidth]{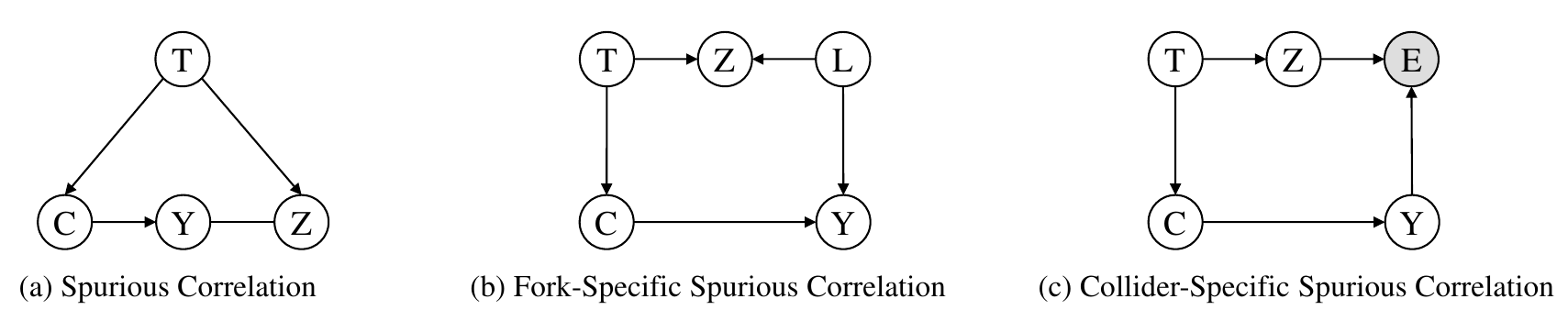}
    \caption{$Y$ and $Z$ form a spurious correlation $Y - Z$ as shown in Figure (a). Figures (b) and (c) model this correlation according to fork-specific spurious correlation and collider-specific spurious correlation, respectively. Although the variable correlations in (b) and (c) are the same, the confounding between $C$ and $Y$ is completely different. Figure (b) is non-confounding, whereas Figure (c) is confounding.}
    \label{fig:example5}
\end{figure}

As illustrated in Figures \ref{fig:example4}(a) and \ref{fig:example5}(a), covariate $Z$ forms spurious correlations with $C$ and $Y$, i.e., in Figure \ref{fig:example4}(a), $Z - C$, and in Figure \ref{fig:example5}(a), $Y - Z$.

Firstly, considering the case of $Z - C$, we cannot analyze the backdoor paths between $C$ and $Y$ using $d$-separation, thus we cannot employ the backdoor criterion to analyze which covariate needs adjustment. Modeling it according to fork-specific spurious correlation and collider-specific spurious correlation, we obtain Figures \ref{fig:example4}(b) and \ref{fig:example4}(c) respectively. In Figure \ref{fig:example4}(b), the backdoor path for $C \rightarrow Y$ is $C \leftarrow L \rightarrow Z \leftarrow T \rightarrow Y$, this path is blocked by the covariate $Z$, therefore, according to Definition ~\ref{do}, $C$ and $Y$ are non-confounding. Although $Z$ is correlated with $C$ through the path $C \leftarrow L \rightarrow Z$ and with $Y$ through the path $Z \leftarrow T \rightarrow Y$, $Z$ cannot act as a confounding factor to interfere with the causal effect of $C \rightarrow Y$.

In Figure \ref{fig:example4}(c), conditioning on $E$ opens the path $C \rightarrow \underline{E} \leftarrow Z \leftarrow T \rightarrow Y$, resulting in confounding between $C$ and $Y$. The correlation between $Z$ and $C$ leads to different confounding scenarios, highlighting the necessity of cautious adjustment for covariate $Z$. In Figure \ref{fig:example4}(b), adjusting $Z$ introduces bias, resulting in:
\begin{equation}
    \sum _{z}\mathbb{P}(Y=y\mid C=c,Z=z)P(Z=z)\neq P(Y=y\mid do(c)).
\end{equation}
Therefore, without clearly understanding the type of correlation denoted by the symbol $-$, one should not arbitrarily use the involved covariates for adjustment. Instead, we perform the adjustment on other variables on the path to control for confounding bias. For instance, in Figures \ref{fig:example4}(b) and \ref{fig:example4}(c), adjusting $T$ yields the correct result:
\begin{equation}
    \sum _{t}\mathbb{P}(Y=y\mid C=c,T=t)P(T=t)= P(Y=y\mid do(c)).
\end{equation}

For the case of $Y - Z$, we model it according to fork-specific spurious correlation and collider-specific spurious correlation, resulting in Figures \ref{fig:example5}(b) and \ref{fig:example5}(c), respectively. In Figures \ref{fig:example5}(b), the backdoor path for $C \rightarrow Y$ is $C \leftarrow T \rightarrow Z \leftarrow L \rightarrow Y$. This path is blocked by the covariate $Z$, so $C$ and $Y$ are non-confounding. In Figure \ref{fig:example5}(c), conditioning on $E$ opens the path $C \leftarrow T \rightarrow Z \rightarrow \underline{E} \leftarrow Y$, resulting in confounding between $C$ and $Y$.

\section{Theoretical Analysis for Robustness of Propensity Score Estimation}
\label{D}
\paragraph{Proof of Corollary 1.}
For any fixed function $\hat{f}$, based on the analysis in Appendix B.2, we consider that when $C$ and $Y$ are non-confounding under the manipulated probability $\mathbb{P}_{m}$, the true risk is defined as $\mathcal{R}(\hat{f})=\mathbb{E}_{(\mathbf{x},y)\in \mathbb{P}_m}\ell(\hat{f}(\mathbf{x}),y)$. The expected risk using the PSW estimator is $\mathcal{R}^{PSW}(\hat{f}|\pi)=\mathbb{E}_{(\mathbf{x},y)\in \mathbb{P}}[\frac{\ell(\hat{f}(\mathbf{x}),y)}{\pi \left ( \mathbf{x} \right )}]$.

Let $v_1, v_2, \ldots, v_n$ be independent random variables such that $a_i = 0 \leq v_i \leq \frac{\ell(\hat{f}(\mathbf{x_i}),y_i)}{\pi \left ( \mathbf{x_i} \right )} = b_i$. We define $S_n = \sum_{i=1}^n v_i$. Then, by Hoeffding's inequality~\cite{hoeffding1994probability}, we have:
\begin{equation}
    \mathbb{P}(\left | S_n-\mathbb{E}[S_n] \right |\geq t)\leq 2e^{-2t^2/\sum (b_i-a_i)^2 }.
\end{equation}

Assume the variable $v_i$ follows a Bernoulli distribution, with $\mathbb{P}(v_i=\frac{\ell(\hat{f}(\mathbf{x_i}),y_i)}{\pi \left ( \mathbf{x_i} \right )})=\pi \left ( \mathbf{x_i} \right )$ and $\mathbb{P}(v_i=0)=1-\pi \left ( \mathbf{x_i} \right )$. Then we have:
\begin{equation}
    \frac{S_n}{n}=\mathcal{R}^{PSW}(\hat{f}|\pi),\quad \mathbb{E}(\frac{S_n}{n})=\mathcal{R}(\hat{f}).
\end{equation}

From equations (20) and (21), we obtain:
\begin{equation}
    \mathbb{P}(\left | \mathcal{R}^{PSW}(\hat{f}|\pi)-\mathcal{R}(\hat{f}) \right |\geq t)\leq 2e^{-2t^2 \cdot n^2/\sum (b_i-a_i)^2 }.
\end{equation}

The above expression is equivalent to:
\begin{equation}
    \mathbb{P}(\left | \mathcal{R}^{PSW}(\hat{f}|\pi)-\mathcal{R}(\hat{f}) \right |\leq t)\geq1- 2e^{-2t^2 \cdot n^2/\sum (b_i-a_i)^2 }.
\end{equation}
Let $\delta' = 2e^{-2t^2 \cdot n^2/\sum (b_i-a_i)^2}$, solving for:
\begin{equation}
    t=\frac{1}{n}\sqrt{\frac{\log\frac{2}{\delta' }}{2}\sum _{i=1}^{n}\frac{\ell^2(\hat{f}(\mathbf{x_i}),y_i)}{\pi^2 \left ( \mathbf{x_i} \right )}}.
\end{equation}

Suppose $\hat{f}_h \in \mathcal{H}$ is an invariant function in the hypothesis space. Let $d_h=\sum_{i=1}^{n}\frac{\ell(\hat{f}_{h}(\mathbf{x_i}),y_i)}{\pi \left ( \mathbf{x_i} \right )}$. Since $\ell(\hat{f}(\mathbf{x}),y)\leq \Omega$, we have $d_h\leq\sum_{i=1}^{n}\frac{\Omega}{\pi \left ( \mathbf{x_i} \right )}$. Combining with equation (22), we obtain:
\begin{equation}
    \sum_{h=1}^{\left | \mathcal{H} \right |}\mathbb{P}(\left | \mathcal{R}^{PSW}(\hat{f}_h|\pi)-\mathcal{R}(\hat{f}_h) \right |\geq t)\leq\left | \mathcal{H} \right |\cdot 2e^{-2t^2 \cdot n^2/\sum \frac{\Omega^2}{\pi^2 \left ( \mathbf{x_i} \right )} }.
\end{equation}

Let
\begin{equation}
    \delta=\left | \mathcal{H} \right |\cdot 2e^{-2t^2 \cdot n^2/\sum \frac{\Omega^2}{\pi^2 \left ( \mathbf{x_i} \right )} }.
\end{equation}
Then, we have:
\begin{align}
&\mathbb{P}(\bigcup_{h=1}^{\left | \mathcal{H} \right |}\left | \mathcal{R}^{PSW}(\hat{f}_h|\pi)-\mathcal{R}(\hat{f}_h) \right |\geq t)\leq \delta \\
\Rightarrow &\mathbb{P}(\max_{\hat{f}_h}\left | \mathcal{R}^{PSW}(\hat{f}_h|\pi)-\mathcal{R}(\hat{f}_h) \right |\leq t)\geq 1-\delta\\
\Rightarrow &\mathbb{P}(\left | \hat{\mathcal{R}}^{PSW}(\hat{f}_h|\pi)-\mathcal{R}(\hat{f}_h) \right |\leq t)\geq 1-\delta.
\end{align}

By solving equation (26), we obtain the generalization error as:
\begin{equation}
    t=\frac{\Omega}{n}\sqrt{\frac{\log\frac{2\left | \mathcal{H} \right |}{\delta }}{2}\sum _{i=1}^{n}\frac{1}{\pi^2 \left ( \mathbf{x_i} \right )}}.
\end{equation}

\section{Experiment Details} \label{sec:experimental_details}
\subsection{Datasets} \label{sec:detail_datasets}
ColoredMNIST is a modified version of the MNIST dataset designed for handwritten digit classification, where each domain within the set \([0.1, 0.3, 0.9]\) features digits spuriously correlated with color. This dataset includes 70,000 samples with dimensions (2, 28, 28) across two classes, distinguishing whether digits are less than 5, with 25\% noise incorporated. RotatedMNIST, another MNIST variant, comprises domains with digits rotated by \(\alpha\) degrees, for \(\alpha \in \{0, 15, 30, 45, 60, 75\}\). It contains 70,000 instances of dimensions (1, 28, 28) classified into 10 categories based on the digit. The PACS dataset encompasses four domains—art, cartoons, photos, and sketches—with 9,991 instances of dimensions (3, 224, 224) spread across seven classes denoting the type of object depicted. VLCS features photographic domains including Caltech101, LabelMe, SUN09, and VOC2007, comprising 10,729 samples with dimensions (3, 224, 224) distributed among five classes indicating the primary object in the photograph. OfficeHome includes four domains: art, clipart, product, and real, containing 15,588 samples with dimensions (3, 224, 224) categorized into 65 classes based on the object type. TerraIncognita, designed for wildlife research, consists of 24,788 photographs taken by camera traps at locations L100, L38, L43, and L46, with dimensions (3, 224, 224) and 10 classes identifying the animal type.

\subsection{Detailed results} \label{sec:detail_results}
All experiments were conducted on Nvidia A100 and Nvidia A800. Here we
report detailed results on each domain of all six datasets. We also use the training domain validation for model selection.

\begin{table}[t]
    \centering
        \caption{ColoredMNIST}
    \begin{tabular}{ccccc}
    \toprule
        Algorithm & +90\% & +80\% & -90\%  & Avg  \\
        \midrule
BalancingERM & 71.5 $\pm$ 0.3 & 71.2 $\pm$ 0.2 & 37.6 $\pm$ 2.9 & 60.1 \\
BalancingCORAL & 70.5 $\pm$ 0.6 & 72.0 $\pm$ 0.2 & 57.2 $\pm$ 3.4 & 66.6 \\
\midrule
BalancingERM + Ours &   71.7 $\pm$ 0.2  &        71.7 $\pm$ 0.3    &      44.2 $\pm$ 7.2    &      62.5  \\
BalancingCORAL + Ours &  71.6 $\pm$ 0.2      &    72.0 $\pm$ 0.2      &    59.7 $\pm$ 4.5       &   67.8 \\
    \bottomrule
    \end{tabular}
    \label{tab:my_label}
\end{table}

\begin{table}[t]
    \centering
        \caption{RotatedMNIST}
   \adjustbox{max width=\textwidth}{ \begin{tabular}{cccccccc}
    \toprule
        Algorithm & 0 & 15 & 30 & 45 & 60 & 75 & Avg  \\
        \midrule
BalancingERM & 94.8 $\pm$ 0.3 & 98.4 $\pm$ 0.1 & 98.7 $\pm$ 0.0 & 98.8 $\pm$ 0.0 & 98.8 $\pm$ 0.0 & 96.4 $\pm$ 0.1 & 97.7 \\
BalancingCORAL & 94.5 $\pm$ 0.4 & 98.7 $\pm$ 0.0 & 98.8 $\pm$ 0.1 & 99.0 $\pm$ 0.0 & 98.9 $\pm$ 0.0 & 96.2 $\pm$ 0.2 & 97.7 \\
\midrule
BalancingERM + Ours & 95.4 $\pm$ 0.2     &     98.8$\pm$ 0.0    &      99.1$\pm$ 0.0      &    99.0 $\pm$0.1      &    99.1 $\pm$ 0.0       &   96.9 $\pm$ 0.1   &       98.1 \\
BalancingCORAL + Ours & 95.2 $\pm$ 0.5   &      98.7 $\pm$0.0      &    98.9 $\pm$ 0.0   &       99.1 $\pm$ 0.1    &      98.8 $\pm$ 0.0    &      96.0 $\pm$ 0.4    &      97.8 \\
    \bottomrule
    \end{tabular}}
    \label{tab:my_label}
\end{table}

\begin{table}[t]
    \centering
        \caption{VLCS}
    \begin{tabular}{cccccc}
    \toprule
        Algorithm & C & L & S & V & Avg  \\
        \midrule
BalancingERM & 96.9 $\pm$ 0.4 & 64.8 $\pm$ 1.2 & 70.2 $\pm$ 0.8 & 72.6 $\pm$ 1.3 & 76.1 \\
BalancingCORAL & 98.3 $\pm$ 0.1 & 63.9 $\pm$ 0.2 & 69.6 $\pm$ 1.1 & 73.7 $\pm$ 1.3 & 76.4 \\
\midrule
BalancingERM + Ours & 96.7 $\pm$ 0.1   &       65.8 $\pm$0.0      &    69.1 $\pm$ 0.0      &    76.8 $\pm$ 0.2       &   77.1  \\
BalancingCORAL + Ours  & 98.0 $\pm$ 0.5      &      65.3 $\pm$ 0.1     &      70.3 $\pm$ 2.0      &     75.8 $\pm$ 0.4     &      77.3 \\
    \bottomrule
    \end{tabular}
    \label{tab:my_label}
\end{table}

\begin{table}[t]
    \centering
        \caption{PACS}
    \begin{tabular}{cccccc}
    \toprule
        Algorithm & A & C & P & S & Avg  \\
        \midrule
BalancingERM & 87.1 $\pm$ 1.8   &       76.7 $\pm$ 2.5      &    97.1 $\pm$ 0.3        &  80.1 $\pm$ 0.4    &      85.2 \\
BalancingCORAL & 87.8 $\pm$ 0.8  & 81.0 $\pm$ 0.1 & 97.1 $\pm$ 0.4 & 81.1 $\pm$ 0.8 & 86.7 \\
\midrule
BalancingERM + Ours & 86.6 $\pm$ 2.4     &     80.6 $\pm$ 1.0      &    97.0 $\pm$ 0.6    &      80.9 $\pm$ 0.6      &    86.3  \\
BalancingCORAL + Ours & 88.6 $\pm$ 0.4     &     81.5 $\pm$ 0.2     &     97.6 $\pm$ 0.5     &     82.1 $\pm$ 1.0   &       87.4 \\
    \bottomrule
    \end{tabular}
    \label{tab:my_label}
\end{table}

\begin{table}[t]
    \centering
        \caption{OfficeHome}
    \begin{tabular}{cccccc}
    \toprule
        Algorithm & A & C & P & R & Avg  \\
        \midrule
BalancingERM & 61.5 $\pm$ 0.4 & 53.8 $\pm$ 0.5 & 75.9 $\pm$ 0.2 & 77.4 $\pm$ 0.5 & 67.1 \\
BalancingCORAL & 65.6 $\pm$ 0.6 & 56.5 $\pm$ 0.6 & 77.6 $\pm$ 0.3 & 78.8 $\pm$ 0.5 & 69.6 \\
\midrule
BalancingERM + Ours & 63.8 $\pm$ 0.8   &   55.9 $\pm$ 0.4      &    75.7 $\pm$ 0.6      &    77.2 $\pm$ 1.0 & 68.2 \\
BalancingCORAL + Ours & 66.0 $\pm$ 0.1 & 57.1 $\pm$ 0.3 & 77.7 $\pm$ 0.5 & 79.5 $\pm$ 0.2 & 70.1 \\
    \bottomrule
    \end{tabular}
    \label{tab:my_label}
\end{table}

\begin{table}[t]
    \centering
        \caption{TerraIncognita}
    \begin{tabular}{cccccc}
    \toprule
        Algorithm & A & C & P & S & Avg  \\
        \midrule
BalancingERM & 53.3 $\pm$ 0.8 & 47.2 $\pm$ 1.9 & 55.3 $\pm$ 0.7 & 36.2 $\pm$ 1.0 & 48.0 \\
BalancingCORAL & 55.2 $\pm$ 0.3 & 42.3 $\pm$ 3.6 & 54.7 $\pm$ 0.4 & 36.0 $\pm$ 1.0 & 47.0 \\
\midrule
BalancingERM + Ours & 57.9 $\pm$ 1.8    &      42.6 $\pm$ 2.7     &     54.8 $\pm$ 1.3   &       37.5 $\pm$ 0.9   &       48.2  \\
BalancingCORAL + Ours & 54.7 $\pm$ 2.1 & 47.8 $\pm$ 2.9    &      51.9 $\pm$ 1.4       &   38.9 $\pm$ 1.6      &    48.3  \\
    \bottomrule
    \end{tabular}
    \label{tab:my_label}
\end{table}

\end{document}